\documentclass[letterpaper, 10 pt, conference]{ieeeconf}  

\IEEEoverridecommandlockouts                              
\usepackage{graphicx} 
\usepackage{amsmath} 
\usepackage{amssymb}  
\usepackage[usenames, dvipsnames]{color}

\title{\LARGE \bf
Incremental Adversarial Learning for Optimal Path Planning
}

\author{Salvatore Virga$^{1}$, Christian Rupprecht$^{1, *}$, Nassir Navab$^{1,2}$ and Christoph Hennersperger$^{1,3}$
\thanks{$^{1}$Salvatore Virga, Christoph Hennersperger, Christian Ruprecht and Nassir Navab are with Computer Aided Medical Procedures, Technical University Munich, Boltzmannstr. 3, 85748 Garching bei Munich, Germany
        {\tt\small salvo.virga@tum.de}}%
\thanks{$^{2}$Nassir Navab is with Johns Hopkins University, Baltmore, USA}
\thanks{$^{3}$Christoph Hennersperger is with Trinity College Dublin, College Green, Dublin 2, Ireland}
\thanks{$^{*}$Christian Rupprecht is now at the University of Oxford, United Kingdom.}
}

\DeclareMathOperator*{\argmin}{arg\,min}

\begin{document}

\maketitle
\thispagestyle{empty}
\pagestyle{empty}

\begin{abstract}
Path planning plays an essential role in many areas of robotics.
Various planning techniques have been presented, either focusing on learning a specific task from demonstrations or retrieving trajectories by optimizing for hand-crafted cost functions which are well defined a priori.

In this work, we present an incremental adversarial learning-based framework that allows inferring implicit behaviour, i.e. the natural characteristic of a set of given trajectories. 
To achieve adversarial learning, a zero-sum game is constructed between a planning algorithm and an adversary - the discriminator.
We employ the discriminator within an optimal motion planning algorithm, such that costs can be learned and optimized iteratively, improving the integration of implicit behavior.
By combining a cost-based planning approach with trained intrinsic behaviour, this can be be integrated also with other constraints such as obstacles or general cost factors within a single planning framework.

We demonstrate the proposed method on a dataset for collision avoidance, as well as for the generation of human-like trajectories from motion capture data.
Our results show that incremental adversarial learning is able to generate paths that reflect the natural implicit behaviour of a dataset, with the ability to improve on performance using iterative learning and generation.
\end{abstract}

\section{INTRODUCTION}
Learning plays an important role not only in robotic motion planning but also in our general lives. 
When children learn new tasks, they are influenced by their peers through demonstrations \cite{hanna1993peer}, but also affected by human anatomy and physiology. 
This interplay results in the development of human behavior, which then is perceived as natural by others.
In robotic motion planning, trajectories that aim to solve a specific task are generated, mainly considering the constraints of the specific robotic system for motion planning.
While some constraints can be easily enforced onto the generated plans to fulfill a given task (e.g. compelling to a workspace), certain criteria remain difficult to specify.
This is, for example, the case for motions that should resemble complex behaviours of a human or animal, and in particular for redundant kinematic chains. Human-like motions can be easier to judge and predict in a human - machine collaborative environment.

\textit{Learning from Demonstration} (LfD) has found many applications over the past years \cite{kober2013reinforcement}, since it avoids hard coding of motion characteristics, but learns them from expert examples. 
The two main approaches in LfD are Behavioural Cloning (BC) and Inverse Reinforcement Learning (IRL). The former relies on supervised learning from state-action pairs obtained by an expert \cite{pomerleau1991efficient}, whereas IRL \cite{ng2000algorithms, abbeel2004apprenticeship} instead aims at learning specific cost functions from the observed behaviours and embedding these into a planning framework.
Here we focus on IRL techniques, as learned behaviors can be easily generalized to tasks that are different from the learned one. This allows to learn an intrinsic behaviour of the demonstrator that is not necessarily specific to the task.
Briefly, IRL approaches are based on two iterative steps: the first computes the optimal policy using the current cost function, while the second updates the cost function itself to maximize (minimize) the next policy.
Techniques based on IRL have been extensively used in robotics, especially for navigation tasks \cite{ratliff2009learning, vasquez2014inverse}. 
These works showed that on the foundation of example paths performed by an expert demonstrator, IRL can be used to find the cost function for which planned navigation paths are optimal.
Recently, both \cite{shiarlis2017rapidly} and \cite{perez2017teaching} focused on learning the cost function used within optimal sampling-based planners for navigation tasks, such as Optimal Rapidly Exploring Random Trees (RRT*) \cite{karaman2011sampling} tasks. 
A focus of these works is the reduction of the computation costs associated with most IRL methods. 
By integrating RRT* for motion planning instead of the classic Markov Decision Process (MDP) schema used for the planning step in most existing IRL models, the authors could show faster navigation planning while providing LfD-like behaviour.

Similar to the concept of Inverse Reinforcement Learning, \textit{Generative Adversarial Networks} (GANs) have recently been presented in unsupervised machine learning to learn a desired cost function \cite{goodfellow2014generative}.
GANs are a class of generative models that consist of two entities -- a generator and a discriminator -- that respectively contest with each other in a game-like fashion.
Up to now, GANs have widely been applied to problems in image generation and domain transfer \cite{goodfellow2014generative,sajjadi2016enhancenet}
In view of robotic applications, \cite{ho2016generative} propose a generic framework that also targets the problem of learning to perform a task from expert demonstrations to overcome the indirect approach used in IRL. 
Rather than iteratively learning a cost function, their approach directly generates distributions of states and actions from the expert data.  
Additionally, \cite{finn2016connection} formally demonstrated the equivalency of GANs to certain IRL algorithms, focusing on sample-base algorithms for maximum entropy.
However, the specific interconnection between IRL and GANs has not yet been much explored, especially in the context of path planning.

In this work, we propose an iterative framework for optimal path planning that takes inspiration from IRL as well as adversarial learning approaches. 
Our work is follows the concept of GANs by using two separate entities for generation and discrimination. 
However, in contrast to GANs, we employ a classical robotic planner to generate paths, which are contesting with a discriminator network trained to differentiate generated trajectories resembling the desired behaviour. 
The advantage of using a planner instead of a neural network is that constraints and guaranties can be easily specified for the planner but are hard to enforce in a neural network.
The aim of the system is to learn a certain motion characteristic from a set of demonstrations. 
Similarly to \cite{ratliff2009learning} and \cite{vasquez2014inverse}, we make use of RRT* for the generation of new trajectories.
RRT* is well-suited for the task at hand as it operates asymptotically optimal, and  can cope with a continuous state space.
In contrast to previous work, we rely on a trained model to discriminate planned trajectories with a desired implicit behaviour instead of optimizing the cost-function directly.
The discriminator is updated iteratively based on trajectories generated by the path planning, generating trajectories mimicking the implicit behaviour of the training data with increasing quality.

We show the proposed iterative optimal architecture for the planning of paths on redundant manipulators rather than focusing on navigation.
This way, we demonstrate how the \emph{implicit behaviour} of demonstrations (e.g. human-like motions) can be integrated with classical planning and cost and local cost estimates for incremental (local) paths.
The planner considers both the implicit behaviour as well as the task-achievement for robotic planning. 
By integrating not only the intended task but also the specific movement characteristics in the plan, one can develop systems that more closely resemble their natural counterparts.
This is of specific interest for robotic trajectory motion planning in redundant manipulators for mimicking human-like behavior.
In medical applications, for example, robotic arms mimicking human motions are likely to result in less resistance or presentiment by patients and staff and thus improved robot-patient interaction.

\section{Learning Implicit Behaviour for Robotic Planning}
In this section, we will describe in depth how we design and couple learning a cost function from examples, with a classical optimal motion planning algorithm, such as RRT*.

The main idea is to instantiate a zero-sum game between the motion planner and an adversary, the discriminator. This adversary is a learned classifier that attempts to determine if a given motion comes from the real data that we are trying to imitate or was generated by the planner. 
The planner, at the same time, has exactly the opposite goal. Its task is to generate motions that \textit{fool} the adversary, i.e. resembling the real data as close as possible.
These opposing goals drive both the planner and the discriminator to improve their performance until planned and real motions become indistinguishable.

\subsection{Data Representation} 
To successfully create a framework that is able to learn a behavioural cost function, specific attention needs to put on a suitable encoding scheme.
This way, it is essential that the encoding is fully agnostic to the underlying, inherent differences between the robot and the sample data that contains the characteristic to be learned. 
For example, when learning to move a robot arm based on real human movements, it is obvious that certain physical properties differ between the human and the robot. 
These can be, for example, arm length, maximal torques, joint limits, or degrees of freedom. Since we are learning a discriminator that tries to decide whether a motion is real or generated, the data representation needs to hide these inherent differences. Otherwise, the classification becomes trivial, and the robot can never improve as it cannot change its physical properties. 

The space of motion representations will be described by $\mathcal{M}$. A robot motion is defined as a sequence of $n$ robot states $M = (s_1, s_2, \ldots, s_n) \in \mathcal{S}^{d\times n}$ in joint space. The representation that exhibits the above-mentioned properties, of hiding the inherent differences between the two domains will be called $r(M) = m: \mathcal{S}^{d\times n} \rightarrow \mathcal{M}$. Since $r$ is hiding details, it is not bijective. Please note that $n$ can be different for every motion, but to avoid cluttering the notation we will not index it further.

\subsection{Adversarial Motion Planning}
\label{sec:motionPlannning}
We formulate the problem as a generic zero-sum game between the planner $P(x) = M: \mathcal{X}\rightarrow\mathcal{S}^{d\times n}$ and the discriminator $D(m): \mathcal{M}\rightarrow [0,1] \subset \mathbb{R}$, where $x \in \mathcal{X}$ represent the constraints (start configuration, goal state, obstacles, etc.) for the motion planner and $m \in \mathcal{M}$ is the representation of a motion. $P$ generates motions $m$ and $D$ assesses their quality by predicting the probability (between $0$ and $1$) that $m$ originates from the desired set $\mathcal{M}^*$. For learning, a set of \textit{real} motions $\mathcal{M}^* \subset \mathcal{M}$ is needed that contains samples from desired motions. These could, for example, be acquired by motion capture of humans. We define the set $\mathcal{M}^P$ as the motions generated by the planner.

Now, the objective for the discriminator $D$ is to identify whether a given motion $m$ comes from the desired set $\mathcal{M}^*$ or has been generated by the planner $P$. Since we model $D$ with a neural network, this resorts to training $D$ with labeled pairs of motions from $\mathcal{M}^*$ as well as $\mathcal{M}^P$. To this end, we parametrize the discriminator $D_{\theta}$ with weights $\theta$ and minimize the following cost function:

\begin{equation}
    \theta^* = \argmin_\theta \sum_{m \in \mathcal{M}^P}\log(D_{\theta}(m)) -\sum_{m \in \mathcal{M}^*}\log(D_{\theta}(m)).
    \label{eq:disc_cost}
\end{equation}

This results in learning $D$ to perform binary classification into generated ($D(m) = 0$) and real ($D(m) = 1$) motions. We can say that $D$ learns to instinctively decide whether a motion is natural or generated. 

The planner has the opposing goal to \textit{fool} the discriminator with its generated motions. Therefore, when we are planning a motion we want to minimize

\begin{equation}
    \log(D_{\theta^*}(r(P(x)))),
    \label{eq:planner_cost}
\end{equation}
thus, minimizing the performance of the discriminator in detecting that this motion actually comes from the planner $P$.

This means that during training we alternate between training the discriminator with the two sets  $\mathcal{M}^*$ and $\mathcal{M}^P$, and generating new motions $\mathcal{M}^P$ with the updated discriminator. 

From theory \cite{goodfellow2014generative} we know that if both $P$ and $D$ have infinite capacity, they are not limited to the possible functions that they can learn (or generate).
On this foundation, the hypothesis is that in the end, the generated motions will be indistinguishable from the desired ones in $\mathcal{M}^*$.

In practice, however, this is only possible if the planner can generate motions sufficiently similar to the desired ones. 
Here, it is once again important to notice why it is crucial that the representation needs to hide inherent (physical) differences between $\mathcal{M}^*$ and $\mathcal{M}^P$.

\subsection{Generating Motions}
While it would be possible to use an additional neural network to generate the motions, we explicitly chose to use a standard state of the art motion planning algorithm in this work. 
This choice is motivated by several considerations. By integrating a traditional motion planning system with the instinct learning framework, all benefits of the former can be inherited. 
This means that adding constraints like obstacles and force limits is readily possible during planning, such that general planning can be performed in conjunction with the trained behaviour.
This is an important differentiation to classic learning from demonstrations, or to the use of a neural network for trajectory planning, as such methods would need to be retrained once these factors change. 
To this end, even if dynamic motions could be planned, it would not be possible to enforce hard constraints if the whole motion is predicted directly. 

While a vast plethora of motion planning algorithms is present in literature, not many allow for an online optimization of the resulting path with respect to a given objective function. 
Our task requires the planner to compute not only a plan that fulfills the motion constraints (start state, goal state, environment, etc.), but rather one that minimizes \eqref{eq:planner_cost} as remarked in the previous section. Therefore, similar to \cite{ratliff2009learning} and \cite{vasquez2014inverse} we make use of a class of planning algorithms commonly defined as \textit{optimal planners}, and in particular of RRT* as state of the art method \cite{karaman2011sampling}.

With the choice of RRT*, a way to incorporate the discriminator $D$ in the planning process is needed.
Only this way, we are able to find motions that the planner classifies as non-generated and thus resembling the \emph{desired behaviour}. 
RRT* explores the space of valid states in a tree structure of robot states $s \in \mathcal{S}$ by selectively expanding nodes $s$ during the search. This means that instead of classifying the whole planned motions as generated or real, we need to estimate the cost $c(s_j) \in \mathbb{R}$ of each node in the tree. 

This leads to the following change in the data representation. While previously we defined $m$ to be the whole planned motion from start to goal, we now augment this, by also including all sub motions from the start state to the intermediate steps of a motion into $\mathcal{M}$. This means that not only is 

\begin{equation}
    r(M) = r(s_1, s_2, \ldots, s_n) \in \mathcal{M},
\end{equation}
but all partial motions starting from $s_1$, too:

\begin{equation}
    r(s_1, \ldots, s_j) \in \mathcal{M},  \forall j = 2, \ldots, n.
\end{equation}

This allows to look at partial, unfinished motions and to classify them. The same has to be done with the target data $\mathcal{M}^*$.

Now, that $D$ also operates on partial motions, we can asses the quality of a partially planned motion by feeding $D$ with the planned motion from the start state to the current node of the state tree. For the planner to operate correctly, we need to compute the cost of a node, and not the cost of the whole motion until then. Naturally, we define the cost $c(s_j)$ of a node $s_j$, as the difference between the quality after adding the node and before. 

\begin{equation}
    c(s_j) = D(r(s_1, \ldots, s_j)) - D(r(s_1, \ldots, s_{j-1}))
\end{equation}

This formulation is well-defined: the sum of costs overall states yields the score of the whole motion. For brevity, we write

\begin{equation}
R(s_j) = r(s_1, \ldots, s_j),
\end{equation}
and yield
\begin{eqnarray}
    \sum_{j=2}^n c(s_j)= & \sum_{j=2}^n [ D(R(s_j)) - D(R(s_{j-1}))]\\ = & D(R(s_n)),
\end{eqnarray}

as every term is subtracted and added once, except the last one, which contains the score of the full motion. The sum starts at $2$ as it needs at least two states to describe a motion. 

Please note that the cost $c$ can be negative in this formulation, as a certain motion can obtain a high score from the discriminator until a \textit{worse} state is reached.
This is a typical behavior for a planned motion; thus $D$ decreases and $c$ becomes negative. 

As detailed in the evaluation, we are employing redundant manipulators in this work. These are considered optimal since their ability to reach a target pose with multiple configurations allows the planner to choose the configuration which optimizes our objective function. Due to this characteristic, multiple goal states are fed to the planning algorithm by uniformly sampling the position of the redundant joint for a given inverse kinematic solution. Only joint positions that are part of a valid joint configuration are taken into account and the respective goal state added to the RRT* tree.

\section{Evaluation}
The goal of this work is to demonstrate the overall concept of iterative trajectory planning using adversarial learning, and we evaluate the described method with two distinct objectives in robotic planning.
The first experiment is to validate that the framework is able to learn desired implicit behaviour from a given sample data.
In a second the objective, our goal is to learn the implicit behaviour of more human-like motions. By using human arm movements collected from a large motion capture dataset, we demonstrate that our method optimizes robotic trajectories to mimic natural human motions.

For all our experiment, a robot model of a 7-DOF anthropomorphic arm has been used. The model was extracted from a complete 32-DOF human body model publicly available~\footnote{\texttt{github.com/baxter-flowers/human\_moveit\_config}}.

\begin{figure}
    \centering
    \includegraphics[width=\linewidth]{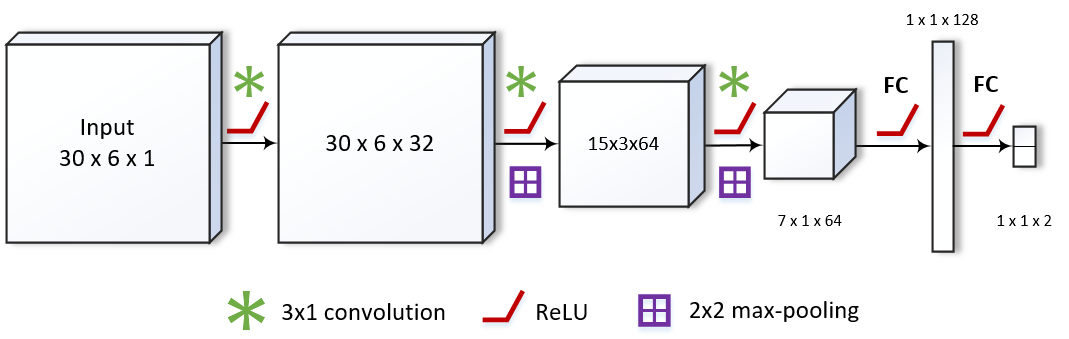}
    \caption{The architecture of the discriminator network.}
    \label{fig:architecture}
\end{figure}

The discriminator consists of a small convolutional neural network; its architecture is depicted in Figure \ref{fig:architecture}. To train and evaluate the network Tensorflow \cite{tensorflow2015-whitepaper} is employed, both in its Python and C++ flavors.
Due to the small size of the input ($30 \times 6 \times 1$) a relatively small network with 3 convolutional layers and one fully connected layer suffices for this task. We train for 10 epochs with an initial learning rate of $0.001$ and use Adam as optimizer.

\begin{figure}
    \centering
    \includegraphics[width=\linewidth]{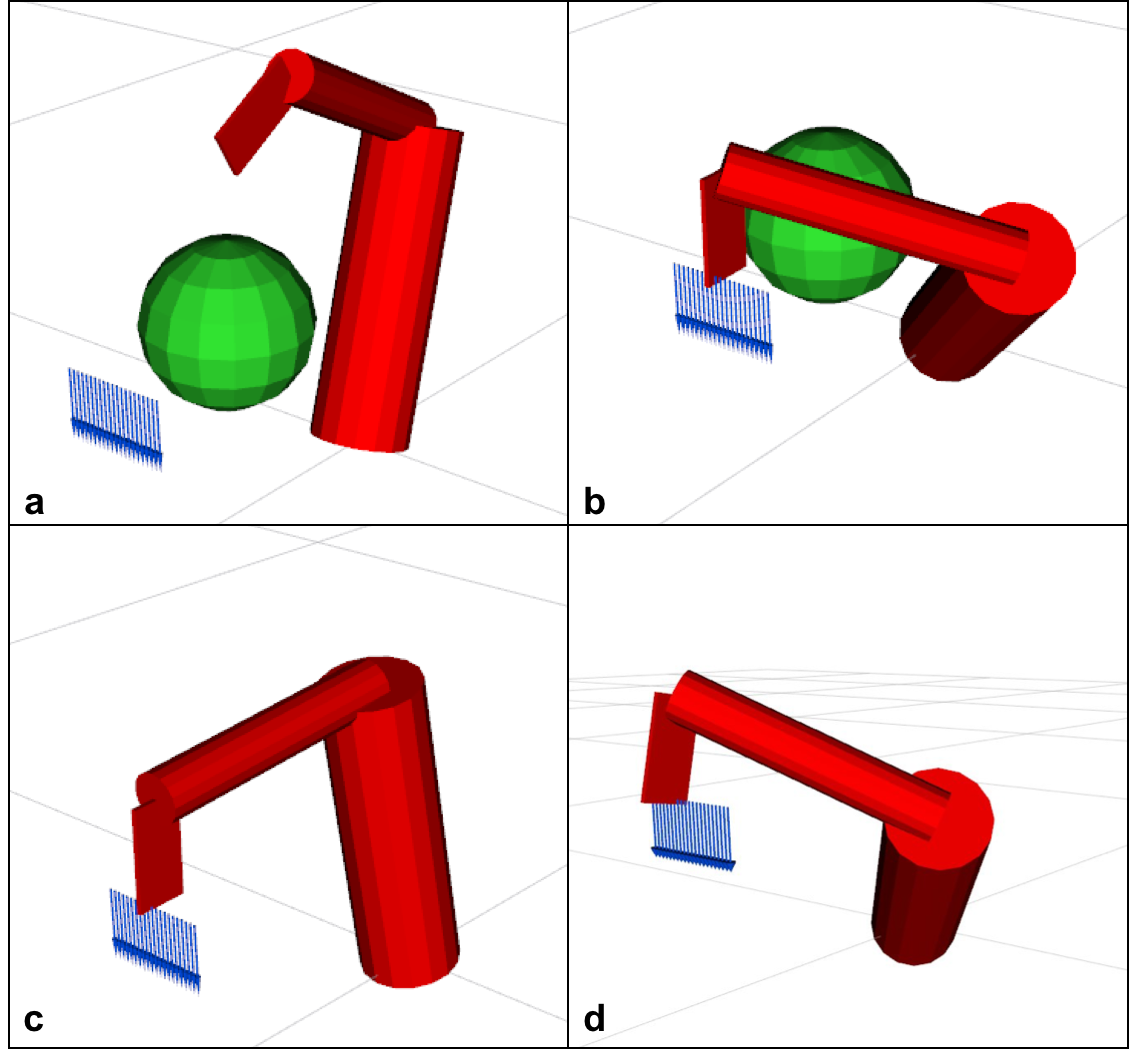}
    \caption{Setup for framework validation. \textbf{a.} 7-DOF Manipulator is displayed in red, the selected goal poses are represented by the arrows in blue and their reach is obstacled by the green sphere. \textbf{b.} Planning trajectories to reach the goal poses taking into account the sphere location result in motions that avoid it. \textbf{c.} Without the occluding sphere, naive planners plan a straight forward trajectory to the goal poses. \textbf{d.} The same planner as c, but using our optimization approach reaches the goal with a similar configuration as b.}
    \label{fig:collision}
\end{figure}

Custom C++ modules integrate the network and motion planning through OMPL~\cite{sucan2012open} and MoveIt!~\cite{chitta2012moveit} have been integrated within the Robot Operating System (ROS) \cite{quigley2009ros}.

\subsection{Learning Implicit Behaviour}
In this first experiment, our goal is to validate that the system can learn a certain implicit behaviour from the sample motion data.  
Since we are interested in the generality of the presented approach, we show that our planning framework can learn from generic motion data.
Beyond that, we also anticipate that it might be difficult to judge the \textit{humanness} in the second experiment.

The experiment is constructed as follows. 
First, we generate the target behavioural data by computing robotic trajectories that reach certain poses in space. 
However, instead of allowing straightforward movements to this point, we place a solid sphere -- as an obstacle -- in front of the targets and add this environment constraint as part of the planning scene, such that the solution provided by the planner algorithm has to reach around the sphere. 
To provide a sufficient data variety, 10 different starting configurations and 21 goal poses, uniformly sampled along a line placed behind the sphere, are selected: This creates 210 trajectories, representing our desired motion data.
That is, we aim at learning the implicit behaviour to avoid the area occupied by the object. The experiment is also represented in Figure \ref{fig:collision}.

To evaluate the intrinsic behavior trained by our method, we then remove the obstacle (sphere) from the scene and plan trajectories for the given start- and end-configurations using RRT* with and without adversarial-planning. 

\begin{figure*}
    \centering
    \includegraphics[width=\textwidth]{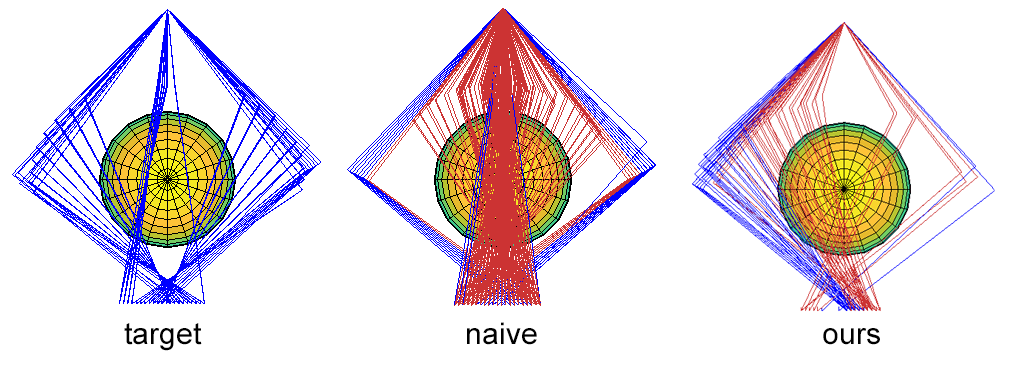}
    \caption{The left figure shows the top view of all robot configurations obtained by planning with the sphere. In the middle, the naive planning without the sphere and on the right with the learned behaviour. Collisions with the sphere are marked in red. Incremental adversarial planning resembles the distribution of the target and is very different from naive planning.}
    \label{fig:collision_distribution}
\end{figure*}

Intuitively, the trained instinct should result in a natural avoidance of region defined by the (invisible) object, while the normal plan would simply consider the lowest cost as minimal distance to the target. 
Quantitatively, the influence can be seen by evaluating the success of planning by means of a planned trajectory not intersecting the invisible object. 
\begin{table}
\centering
\begin{tabular}{l|rrrr}
 & naive & iteration 1 & iteration 2 & iteration 3 \\ \hline
 success & 12.70\% & 22.16\% & 25.93\% & \textbf{36.25\%} \\
\end{tabular}
\caption{\label{tab:collision_results}Success rate, in learning to avoid an invisible object just from examples. After only three iterations, the performance has nearly tripled.}
\end{table}
We compute how many planned trajectories intersect the sphere. The better we can learn to imitate the target data. the less collisions we should cause.
As it can be seen in Table \ref{tab:collision_results}, planning without the obstacle and our planner has a success rate of $12.7\%$, which means that in $87.3\%$ cases the planner hits the invisible object. This is expected since it is simply not aware of any obstacles in the scene.
After the second and third iteration, the planner has learned to implicitly avoid the invisible object 26\% and 36\% of the time. For the remaining trajectories that do collide, the general configuration is still correct. Most of the time the trajectory only slightly clips the sphere shortly during movement. 

A visualization of the end pose of the robot for different starting and goal states is given in Figure \ref{fig:collision_distribution}.
In the target scenario, the planning is performed by adding the sphere as an obstacle, thus the final states have the robot reach around the sphere. Naive planning without knowledge of the sphere directly moves towards the goal. Thus almost all configurations accumulate on top of the sphere but intersect. After training, our planner has managed to recreate a similar distribution as the target, without planning configurations which end on top of the sphere. Still, some configurations intersect, but the overall configuration is correct and reflects the characteristics of the target plans. A bias can be seen towards the left. Since the starting configurations were not chosen symmetrically, there was a bias in the target data towards the left which the model then learns.

With this experiment, we are able to show that the proposed system can infer and transfer instincts from the target motion set. The planner uses the implicit behaviour in the form of the learned discriminator that is then used to compute the cost of each node during planning. 

\subsection{Learning to be Human}
For the real data, we download the CMU Graphics Lab Motion Capture (MOCAP) Database\ (http://mocap.cs.cmu.edu) and parse it to extract arm motions only. Every motion contains the shoulder, elbow and hand joint, with the shoulder considered as the base of the kinematic chain.
Due to the general scope of such large dataset, the data undergoes various pre-processing steps:
\begin{itemize}
\item \textit{Direction normalization} - A traditional planner expects the robot description to have a fixed base, thus partially limiting its workspace. The data from the MOCAP Database contain arm motions that overall span a much larger workspace. Each arm trajectory in the dataset is rotate around its base such that the mean hand directional  $\vec{d}_{\,hand}$ would align to the Z-axis vector $(0,0,1)_z$.

\item \textit{Velocity Splits} - Trajectories extracted from the MOCAP dataset have arbitrary duration and path. Using motions that include loops or long and more chaotic paths would undermine the quality of the generated motions. To obtain a more smooth initial dataset, the trajectories are split into smaller segments accordingly to the hand mean velocity, such that segments have $\vec{v}_{\,hand} \geq \tau_v$, with $\vec{v}_{\,hand}$ the mean hand velocity and $\tau_v$ a velocity threshold. That is, selecting a low enough threshold value, we split the trajectories using the samples that correspond to a single motion being complete within the overall sequence.
\end{itemize}

This data is then used to create the target set $\mathcal{M}^*$ by using the transfer function $r$. In this case, we resample all trajectories to 30 points and save the 3D location of all joints. These joints are then normalized to joint directions instead of joint segments. This grants a certain invariance to the actual length of a segment since it only encodes direction.
In total we generate $50,563$ samples from the motion capture database, which are then used to train \emph{humanness} according to Sec. \ref{sec:motionPlannning}. One iteration of planning and learning takes roughly 2 hours, while most of the time is spent generating paths. Training the small network only takes about 10 minutes on a NVIDIA TitanXp GPU.

\begin{table}
\centering
\begin{tabular}{l|rrrrr}
 RMSE (meters) & naive & iter. 1 & iter. 2 & iter. 5  & iter. 10\\ \hline
 elbow & 0.333 & 0.317 & 0.299 & 0.277 & \textbf{0.272} \\
 hand & 0.005 & \textbf{0.001} & \textbf{0.001} & \textbf{0.001} & \textbf{0.001} \\
\end{tabular}
\caption{\label{tab:human_results}Root Mean Squared Error in meters between the human motion capture and our method after some iterations.}
\end{table}

One way to assess the humanness quantitatively is to try to repeat human motions. Thus, we use 600 unseen motions from the motion capture data and use start and end configuration as start and goal state for the planner.
Table \ref{tab:human_results} shows the root mean squared error between the human target and the planned trajectories. With our incremental adversarial planning the error reduces compared to the direct motion planning. This means that the created trajectories become closer to human ones in Cartesian space.

\section{Discussion}
The results from our first experiment show that it is possible to learn implicit behaviour using incremental adversarial learning but also integrate additional constraints to the planning, such that flexible and versatile paths can be planned dynamically.
Considering the specific aim of mimicking human \emph{behaviour}, our results show that planning using instinct can provide realistic human-like trajectories for dynamic pose configurations. Although the original dataset was not handcrafted for this task; a more specific task created to capture a specific \textit{implicit behaviour} (e.g. motion capture of a human arm grasping objects), could lead to better performances. We also observe that iterating between learning and planning improves performance step by step.

The combination of learned (intrinsic) behavior with classic planning approaches presents a promising way forward for robotic path optimization.
Not only can specific regions be avoided or preferred, but general intrinsic behavior can be effectively integrated with other factors (hard constraints, obstacles) for path planning. 

To this end, we believe that the present framework can also be used in other areas than robotic path planning.
One example would be traditional planning problems such as path-finding in a maze, where one could learn specific instincts. A scared actor would slow down before corners or walk alongside the wall instead of the middle of the corridor. 
In overall, for mimicking human facial expressions, a learned humanness score could also be very valuable in generating believable emotions. 

In moving forward, our ongoing work is focused on the continued generalization of the approach presented in this work. 
Specifically, our interest lies in the evaluation of the extent to which it is possible to use the same learned instinct model for different robot configurations such as different 7-DOF systems. 
By integrating a more diverse set of sample instinct data, the aim is to demonstrate general, natural behavior for a certain group of robot systems.
\section{Conclusions}
We have presented a method that is able to learn motion planning from arbitrary examples and tries to imitate them, while still adhering to typical motion planning constraints.
The framework is presenting an efforts towards incremental adversarial learning to learn implicit behaviour from a dataset and then apply it during planning. This opens up many exciting research directions for future work. 


\section*{ACKNOWLEDGMENT}

This work partially received funding by the project IOTMA supported by the central innovation program for SMEs (ZIM), the BayMED project 5G-MedServices funded by the Bavarian state, as well as from the European Union’s Horizon 2020 research and innovation program EDEN2020 under Grant Agreement No. 688279. The motion capture data used in this project was obtained from mocap.cs.cmu.edu which was created with funding from NSF EIA-0196217. We gratefully acknowledge the support of NVIDIA Corporation with the donation of the Titan X Pascal GPU used here. 

\bibliographystyle{IEEEtran}
\bibliography{IEEEabrv,references}

\end{document}